\def\cvprPaperID{0000}
\def\confName{CVPR}
\def\confYear{2023}

\def\authorBlock{
    Kairui Yang$^{1}$\footnotemark[1] \enskip \enskip 
    Enhui Ma$^{1}$\footnotemark[1] \enskip\enskip 
    Jibin Peng$^{1}$ \enskip\enskip 
    Qing Guo$^{2}$ \enskip\enskip
    Jianping Wu$^{3}$ \enskip\enskip
    Di Lin$^{1}$\footnotemark[2] \enskip\enskip
    Kaicheng Yu$^{4}$\\
    \small
    \textsuperscript{1}Tianjin University \enskip
    \textsuperscript{2}IHPC and CFAR, Agency for Science, Technology and Research, Singapore \enskip 
    \textsuperscript{3}Tsinghua University \enskip 
    \textsuperscript{4}Westlake University \enskip \\
}

\newif\ifreview 
\newif\ifarxiv \newcommand{\arxiv}{\arxivtrue}
\newif\ifcamera 
\newif\ifrebuttal

\arxiv % \review OR \arxiv OR \cameraready

\pdfoutput=1
\documentclass[10pt,twocolumn,letterpaper]{article}

\usepackage{threeparttable}
\usepackage{times}
\usepackage{epsfig}
\usepackage{graphicx}
\usepackage{amsmath}
\usepackage{mathrsfs}
\usepackage{amssymb}
\usepackage{tabulary}
\usepackage{color}
\usepackage{multirow}
\usepackage{bm}
\usepackage{makecell}
\usepackage{bbding}
\usepackage{threeparttable}
\usepackage{color}

\usepackage{booktabs}
\ifreview \usepackage[review]{cvpr} \fi
\ifarxiv \usepackage[pagenumbers]{cvpr} \fi
\ifrebuttal \usepackage[rebuttal]{cvpr} \fi
\ifcamera \usepackage{cvpr} \fi

\usepackage{graphicx}
\usepackage{amsmath}
\usepackage{amssymb}
\usepackage{booktabs}

\usepackage{times}
\usepackage{microtype}
\usepackage{epsfig}
\usepackage[table,xcdraw]{xcolor}
\usepackage{caption}
\usepackage{float}
\usepackage{placeins}
\usepackage{color, colortbl}
\usepackage{stfloats}
\usepackage{enumitem}
\usepackage{tabularx}
\usepackage{xstring}
\usepackage{multirow}
\usepackage{xspace}
\usepackage{subcaption}
\usepackage{xcolor}
\usepackage[hang,flushmargin]{footmisc}

\ifcamera \usepackage[accsupp]{axessibility} \fi

% ...

\ifarxiv  \fi

\newcommand{\R}[1]{{%
    \textbf{%
        \ifstrequal{#1}{1}{\textcolor{red}{R#1}}{%
        \ifstrequal{#1}{2}{\textcolor{blue}{R#1}}{%
        \ifstrequal{#1}{3}{\textcolor{magenta}{R#1}}{%
        \ifstrequal{#1}{4}{\textcolor{teal}{R#1}}{%
                           \textcolor{cyan}{R#1}%
        }}}}%
    }%
}}

\usepackage{xr-hyper}

\makeatletter
\newcommand*{\addFileDependency}[1]{
  \typeout{(#1)}
  \@addtofilelist{#1}
  \IfFileExists{#1}{}{\typeout{No file #1.}}
}

\makeatother

\usepackage[pagebackref,breaklinks,colorlinks]{hyperref}
\usepackage[capitalize]{cleveref}
\crefname{section}{Sec.}{Secs.}
\crefname{table}{Table}{Tables}
\crefname{figure}{Fig.}{Figs.}

\begin{document}
%% TITLE
\title{BEVControl: Accurately Controlling Street-view Elements with\\Multi-perspective Consistency via BEV Sketch Layout}
\author{\authorBlock}
% \maketitle
\twocolumn[{%
	\maketitle
	\renewcommand\twocolumn[1][]{#1}%
	\begin{center}
		\centering
		% \includegraphics[width=0.99\textwidth]{fig/teaser_v2.pdf}
            % \vspace{-0.25in}
            \includegraphics[width=1.0\textwidth]{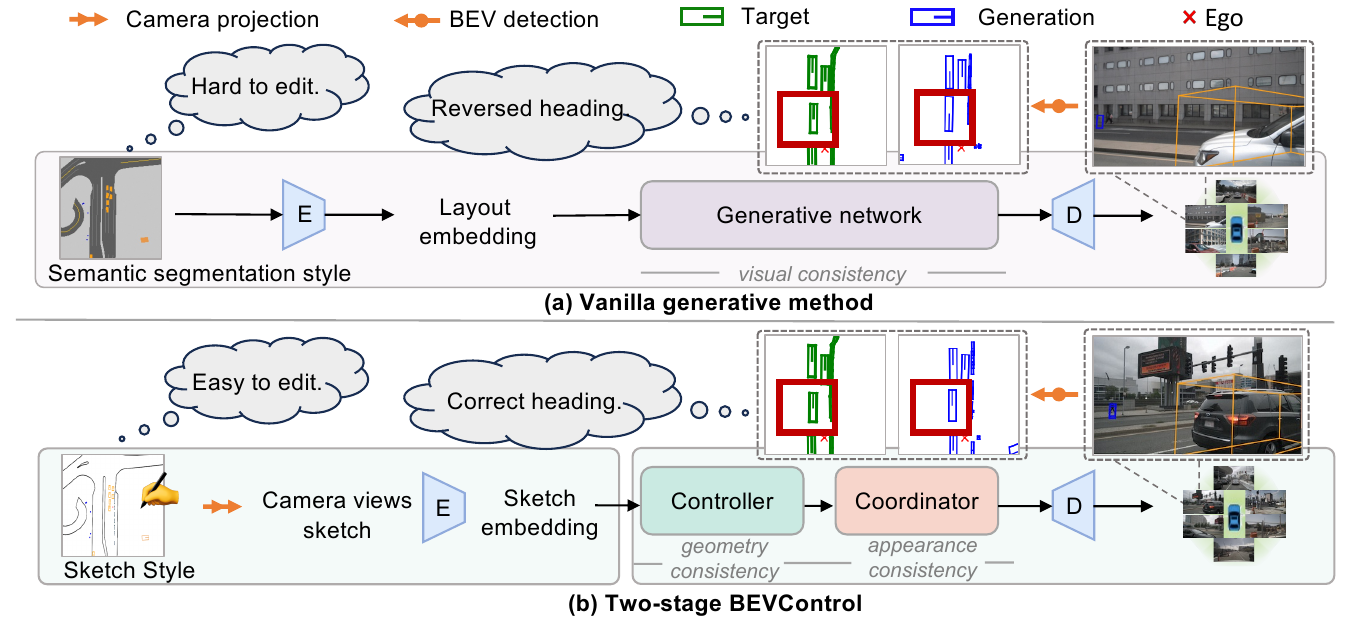}
            % \vspace{-0.25in}
            \vspace{-0.25in}
		\captionof{figure}{
        \textbf{Comparison between different generative networks hinted by Bird's Eye View~(BEV) segmentation layout v.s. sketch layout.} \textbf{(a)} Vanilla generative pipeline feeds a semantic segmentation style input into a generative network and outputs reasonable multi-view images. However, we discover that it fails to generate accurate object-level details. For example, we show a common failure of a state-of-the-art algorithm where the generated vehicle has reversed heading compared to the target 3D bounding box. In addition, editing the semantic segmentation style input is a hard task and requires non-trivial human effort. \textbf{(b)} To this end, we propose a two-stage method that provides finer background and foreground geometry control, dubbed BEVControl. It supports sketch style input that enables fast and easy editing. In addition, our BEVControl decouples visual consistency into two sub-goals: achieving geometry consistency between street and bird's-eye views through the Controller; and achieving appearance consistency between street views through the Coordinator. }
		\label{fig:teaser}
	\end{center}
}

]

\renewcommand{\thefootnote}{\fnsymbol{footnote}}
\footnotetext[1]{Co-first authors.}
\footnotetext[2]{Corresponding author.}
% \footnotetext[3]{Work done during the internship at Alibaba Damo Academy.}
\renewcommand{\thefootnote}{\arabic{footnote}}

\begin{abstract}
\vspace{-10pt}

% BEV perception allows the multi-perspective image information to assist the autonomous driving cars, which are empowered to sense the street-view contents in a wide area exquisitely. Yet, how to collect substantial data of BEV layouts and the corresponding street-view images for training the BEV perception model needs to be addressed. 
% % \KY{still lacks a reasoning from collection to lacking data to need of generation.}
% % Hinted 
% \ky{Guided} by the BEV segmentation layouts, the 
% latest generative networks seem to synthesize \ky{photo-realistic} street-view images 

Using synthesized images to boost the performance of perception models is a long-standing research challenge in computer vision. It becomes more eminent in visual-centric autonomous driving systems with multi-view cameras as some long-tail scenarios can never be collected. 
Guided by the BEV segmentation layouts, the existing generative networks seem to synthesize photo-realistic street-view images when evaluated solely on scene-level metrics. However, once zoom-in, they usually fail to produce accurate foreground and background details such as heading. 
To this end, we propose a two-stage generative method, dubbed BEVControl, that can generate accurate foreground and background contents. In contrast to segmentation-like input, it also supports sketch style input, which is more flexible for humans to edit. In addition, we propose a comprehensive multi-level evaluation protocol to fairly compare the quality of the generated scene, foreground object, and background geometry. Our extensive experiments show that our BEVControl surpasses the state-of-the-art method, BEVGen, by a significant margin, from 5.89 to 26.80 on foreground segmentation mIoU. In addition, we show that using images generated by BEVControl to train the downstream perception model, it achieves on average 1.29 improvement in NDS score.

% Code will be made publicly available at XXX.
% with a realistic style. 
% Yet
% \ky{However}, they lack control of the specific details of the synthesized image contents. 
% In this paper, we introduce BEVControl potent for controlling the background and foreground contents, thus enabling a flexible scheme to enrich the training data for BEV perception. BEVControl regards the BEV sketch layout as the hints for accurately controlling the appearances of the background and foreground elements in the generated street-view images. We can easily edit the BEV sketch layout. The edited counterparts can be used by BEVControl to produce richer data of the street-view images. Furthermore, BEVControl has the cross-view-cross-element attention, which preserves the visual consistency across multiple perspectives of street-view contents. 
% The extensive experiments demonstrate the strong controlling power of BEVControl. 
% The BEV sketch layouts and the corresponding street-view images newly yielded by BEVControl augment the training of the BEV perception model, eventually improving the performances of object recognition from the bird's-eye view.
\vspace{-20pt}
\end{abstract}
\section{Introduction}
\label{sec:intro}

BEV perception for autonomous driving has become popular. It requires understanding the objects in the streets captured from multiple cameras' views, where the things should correspond to the positions from the bird's-eye perspective. The street and bird's-eye views allow the autonomous driving car to broadly sense the objects, thus advancing the progress on an array of downstream applications (e.g., street-view object recognition~\cite{li2022_bevformer, zhou2022_CVT} and traffic flow prediction~\cite{hu2021_fiery, akan2022stretchbev, zhang2022_beverse}).

In today's age of deep learning, reliable BEV perception heavily relies on deep networks trained on many street-view images and the corresponding BEV segmentation layouts, to enable the autonomous car's self-control. To achieve large-scale data for BEV perception, someone may employ autonomous vehicles to travel around the city while recording the street-view images by multiple cameras and mapping the objects to the BEV segmentation layout. Undoubtedly, this solution reduces the human effort for data collection. Yet, autonomous cars without perfect self-control may give rise to traffic congestion or even fatal accident. Moreover, someone must annotate objects across the street and bird's-eye views at an expensive cost. Extra effort is needed to double-check the consistency of annotation across various views.

% Rather than laboriously collecting street-view images from the natural environment and annotating multi-view photos, there are many works~\cite{?} borrowing the success of the fast-growing family of generative networks~\cite{?} for creating new street-view images with a realistic style, which augment the training data for BEV perception. Typically, these methods feed the BEV layout into the generative network. The BEV layout provides the semantic categories and spatial distribution of the objects in the street for the generative network, thus controlling the content of the generated images. Even with an identical BEV layout, the generative network can randomly associate diverse appearances to the objects already appearing in the street. The BEV layout is editable, allowing new things to be added to the generated street-view images. The existing methods generally focus on improving the visual consistency between various street views and the geometric correspondence between the street and bird's-eye views. Nevertheless, only reasonable visual consistency and geometric correspondence are inadequate for the data augmentation, which also requires a diversity of visual elements (e.g., weather condition, road layout, lane line, and vehicle position/heading) in the street-view images to enhance the generalization power of the BEV perception models. The generative network should accurately control the visual elements to achieve data diversity - this is cared less by the existing methods.

Rather than laboriously collecting street-view images from the natural environment and annotating multi-view photos, there are many works~\cite{swerdlow2023_bevgen} resort to the fast-growing family of generative networks~\cite{van2017_vqvae, chang2022_maskgit} for creating new street-view images with a realistic style, which augment the training data for BEV perception. As illustrated in Figure~\ref{fig:teaser}(a), these methods feed the BEV layout into the generative network. The BEV segmentation layout provides the semantic categories and spatial distribution of the objects in the street for the generative network, thus controlling the content of the generated images. Even with an identical BEV layout, the generative network can randomly associate diverse appearances to the objects already appearing in the street.

In spite of the success of the generative networks, the current methods consider less about two critical issues when generating street-view images based on the BEV segmentation layouts. First, the BEV segmentation layout can be analogy to the panoptic segmentation map, where the background stuff and foreground objects unanimously have the pixel-wise annotations in details. It is inconvenient to edit the details of the BEV segmentation layout, thus disallowing many layouts with diversity to be produced for further enriching the street-view data. Second, the existing methods generally focus on improving the visual consistency between various street views and the geometric correspondence between the street and bird's-eye views. Nevertheless, only reasonable visual consistency and geometric correspondence are inadequate for the data augmentation, which also requires a diversity of visual elements (e.g., road layout, lane line, and vehicle position/heading) in the street-view images to enhance the generalization power of the BEV perception models. For this purpose, the generative network should accurately control the visual elements to achieve data diversity.

\begin{figure*}[t!]
\centering
\includegraphics[width=\linewidth]{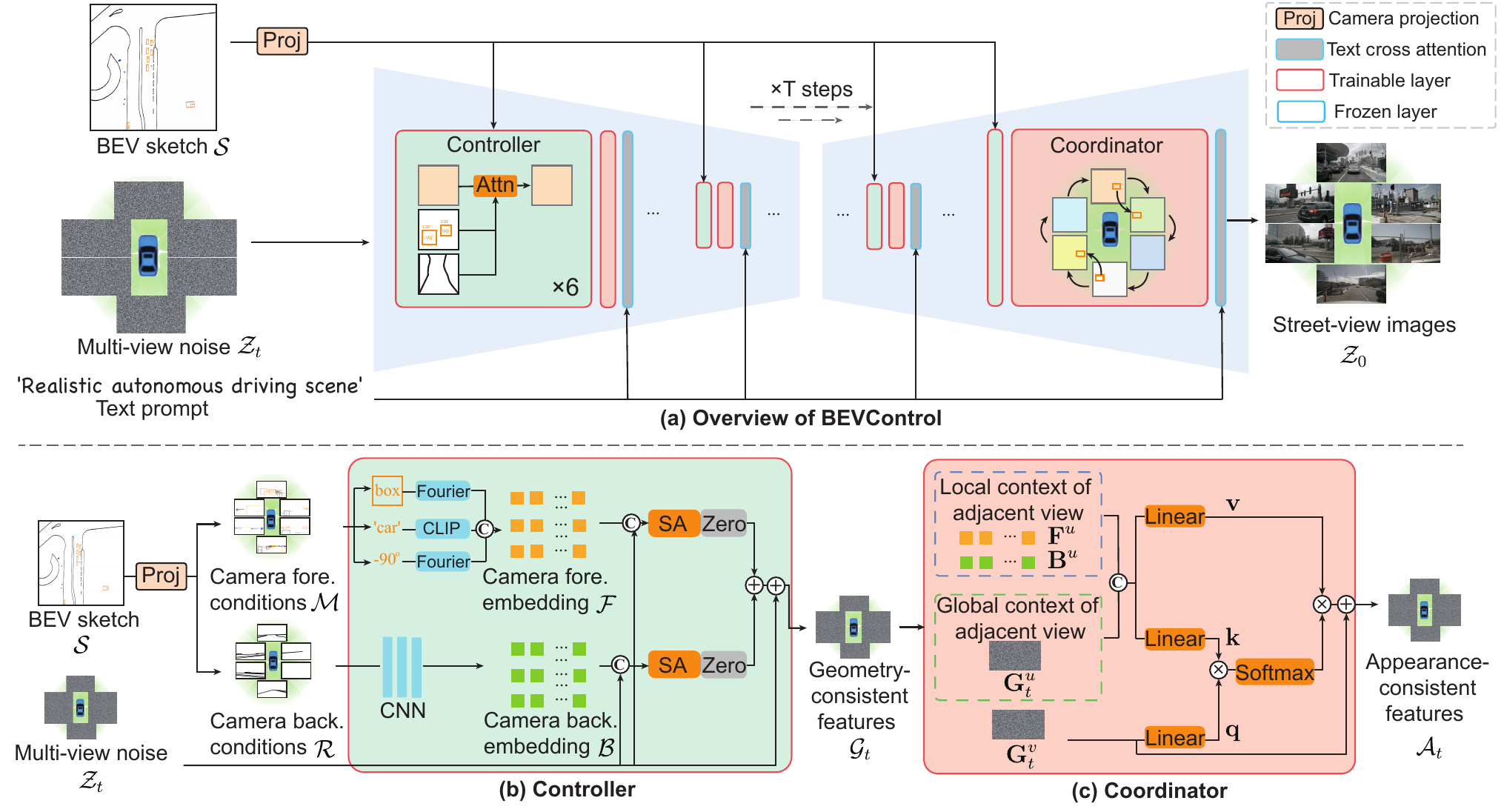}
\vspace{-0.1in}
\caption{\textbf{(a) Overview of BEVControl.} 
It takes inputs as an edit-friendly BEV sketch $\mathcal{S}$, multi-view noisy images $\mathcal{Z}_t $ and text prompt, generating multi-view images $ \mathcal{Z}_0 $. BEVControl is a UNet structure generative network composed of a sequence of modules. Each module has two elements, controller and coordinator. Each controller takes input from BEV sketch features extracted from the projection module. See Fig.~\ref{fig:proj_process} for more details. Text features are encoded cross-attention as in \cite{rombach2022_ldm}. \textbf{(b) Details of Controller.}  A controller module takes in the foreground and background location information of the camera views sketch in a self-attention manner and outputs
the geometry-consistent street view features $\mathcal{G}_t$ concerning the BEV sketch $\mathcal{S}$. \textbf{(c) Details of Coordinator.} A coordinator module leverages a novel cross-view-cross-element attention mechanism that enables context interaction across views, outputting the appearance-consistent street view features $\mathcal{A}_t$. }
\label{fig:overview}
\vspace{0.2in}
\end{figure*}

This paper proposes BEVControl, which has a strong power for controlling the visual elements of the generated street-view images based on the BEV sketch layout. We illustrate the architecture of BEVControl in Figure~\ref{fig:teaser}(b). BEVControl has the controller and coordinator. The controller relies on the sketches of the background (e.g., road layout and lane line) and foreground elements (e.g., vehicle and pedestrian), which are easier to be edited than the pixel-wise annotations on the segmentation layout, to control the appearances and geometric properties of these two kinds of elements in the generated street-view images separately. The coordinator attends to the underlying correlation between the background and foreground elements, whose visual consistency across different views is preserved.

The controller regards the background and foreground elements' sketches as hints. Here, the sketch and bounding boxes mainly represent the geometric shapes of the background and foreground elements. They are mapped from the identical BEV sketch layout, thus preserving the geometric correspondence between the elements across the street and bird's-eye views. With the hints attending to the background and foreground elements respectively, the controller employs the diffusion model to compute the latent feature maps of the street-view images, which represent various perspectives captured by multiple vehicle cameras. We feed these street-view feature maps to the coordinator. The coordinator uses a novel cross-view-cross-element attention mechanism to comprehensively model the context of visual elements in different views. It uses the context to enhance the visual consistency between the visual elements from multiple street perspectives, eventually producing street-view images.

We extract the BEV sketches from the public dataset, nuScenes~\cite{caesar2020_nuscenes}, to drive BEVControl to generate the street-view images for the classical object detection task. In contrast to the current methods that primarily mind the usefulness of the generated data for improving the performances on down-stream tasks, we extensively evaluate the controlling power of BEVControl, which helps to yield richer training data and achieve state-of-the-art object detection performance on nuScenes. We brief our contributions below:
\begin{itemize}
\item We use the cost-effective BEV sketch layouts to easier produce a large amount of street-view images.
\item We propose the sketch-based BEVControl, which has a strong control of the background and foreground elements in the generated street-view images.
\item BEVControl remarkably augments the training dataset, which helps to achieve state-of-the-art object detection results on nuScenes.
\end{itemize}

\section{Related Work}
% \label{sec:related}

The literature on image generation is vast~\cite{rombach2022_ldm,tang2023_mvdiffusion}. We mainly survey the approaches to the conditional generation of images with visual consistency. These approaches are closely relevant to our work because they also leverage various types of image information to control the image contents.

\vspace{0.05in}
\noindent{\bf Image Generation via Multi-modal Information~~}
The recent progress in image generation is primarily attributed to the generative networks pre-trained on large-scale image data. Amidst a broad range of generative networks, the family of diffusion models~\cite{rombach2022_ldm, nichol2021_glide, ramesh2022_hierarchical, saharia2022_photorealistic} lead a fashion of using multi-modal information for generating the image contents. The latent diffusion model~\cite{rombach2022_ldm} is a framework for producing images based on text. Note that the text-based information roughly specifies the image contents, disallowing a fine-grained control of the image contents. To address the above problem, the multi-modal image information like layout images~\cite{li2023_gligen, yang2023_reco, zheng2023_layoutdiffusion, cheng2023_layoutdiffuse}, semantic segmentation maps~\cite{mou2023_T2i_adapter, xue2023_freestyle_Layout_to_Image, avrahami2023_spatext, ham2023_modulating, wang2022_semantic, bar2023_multidiffusion}, object sketches~\cite{mou2023_T2i_adapter, huang2023_composer, bashkirova2023_masksketch, voynov2022_sketch, zhang2023_controlnet}, and depth images~\cite{huang2023_composer, zhang2023_controlnet} have been involved for hinting the image generation.

Generally, the above methods concentrate on generating a single image, where the image contents' semantic categories are aligned with the hints. This paper considers a more complex setting where the street-view image contents of multiple perspectives are generated. In addition to the semantic categories, we should accurately control the geometric properties of the generated street-view images. This goal is non-trivial, especially when the geometric patterns of the foreground and background contents are diverse. To achieve this goal, we resort to the appropriate modalities for controlling the foreground and background contents, thus enhancing the controlling power of the generative network.

\vspace{0.05in}
\noindent{\bf Multi-view Image Generation with Visual Consistency~~}
The visual consistency is a natural property of the authentic images of multiple views. Similarly, we should preserve the visual consistency across multi-view images generated by the deep network. For this purpose, MVDiffusion~\cite{tang2023_mvdiffusion} uses the cross-view attention mechanism to create panoramic images from text, maintaining the global correspondence of multi-view images. The video generation methods~\cite{khachatryan2023_text2video_zero, wu2022_Tune_A_Video, zhang2023_controlvideo, chu2023_video_ControlNet} use the temporal cross-frame attention to preserve the visual consistency across distinct views of image contents at different moments. BEVGen~\cite{swerdlow2023_bevgen} is a contemporary work that generates multi-view images of the street based on the BEV segmentation layout. It employs an auto-regressive transformer with cross-view attention to maintain visual consistency across multi-view images.

The above methods usually work well when the global appearances of multi-view images coincide. But they are less effective for preserving the multi-view consistency when more accurate control of the individual contents (e.g., the orientations of different cars) is desired. This is because the independent operations of content control easily lead to inconsistency across other contents in the same venue. In contrast to the existing methods, we propose cross-view cross-object attention, which remarkably augments the visual consistency of the generated multi-view images.

\section{BEVControl}
\label{sec:method}

We illustrate the overall architecture of BEVControl in Figure~\ref{fig:overview}. 
Following the LDM~\cite{rombach2022_ldm}, BEVControl is a classic UNet structure consisting of an encoder and a decoder. They are composed of three modules stacked multiple times: controller, coordinator, and text cross-attention. 
We process all image features in the latent space, so the image features below specifically refer to those in the latent space.

At first, BEVControl takes the edit-friendly BEV sketch $\mathcal{S} \in \mathbb{R}^{K \times K \times 5}$, text description, and street-view noisy images  $\mathcal{Z}_t = \{ \textbf{Z}_t^v \in \mathbb{R}^{ H \times W \times C} ~|~ v=1,...,V \}$  as input. Here, $V$ denotes the number of perspective views. 
% To simplify the symbol representation,
All sets denoted by $\{\cdot\}$ represent $V$ viewpoints in the method below to foster better readability.
$\mathcal{S}$ is an editable canvas, which supports editing the objects within a 160 $\times$ 160-meter range around the ego car. The five channels of $S$ represent the background sketch (road line), pixel coordinates of the box's center point, text label, and heading of foreground objects, respectively.
In training, $\mathcal{Z}_t$ is a noisy version of street-view authentic images $\mathcal{Z}_0$ by forward diffusion process of ~\cite{rombach2022_ldm}. In inference, $\mathcal{Z}_t$ is street-view noise sampled from $ \mathcal{N}(0, \textbf{I}) $. 
$H, W$, and $C$ expressly represent the spatial resolutions and channels of latent features.

% proj process
\begin{figure}[t!]
\centering
\includegraphics[width=\linewidth]{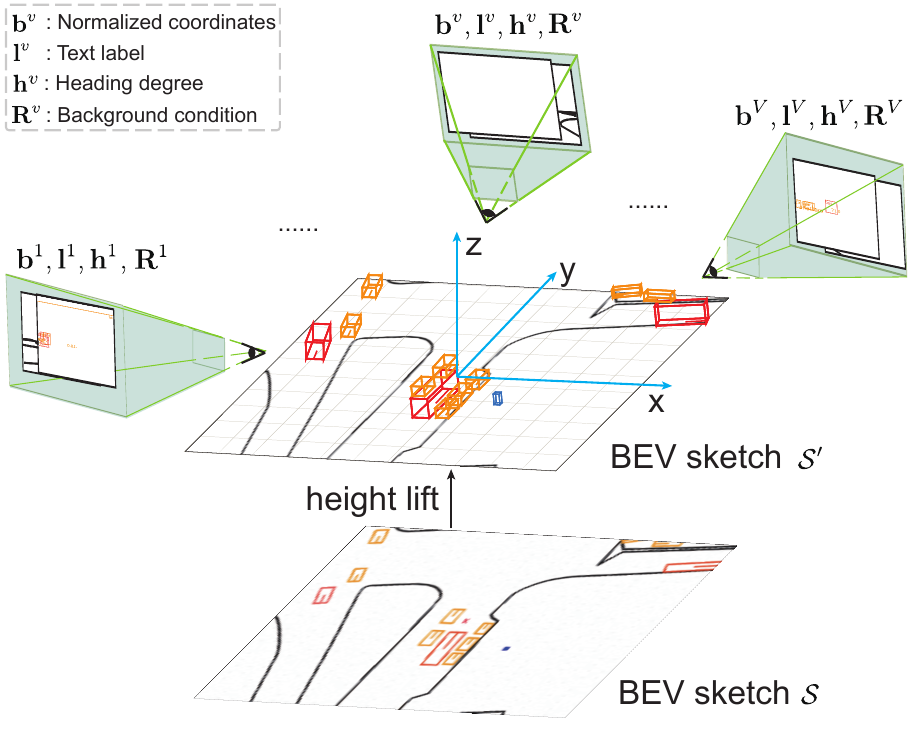}
\vspace{-0.1in}
\caption{The camera projection process from BEV sketch. }
\label{fig:proj_process}
\vspace{-0.15in}
\end{figure}

BEVControl first projects the BEV sketch $\mathcal{S}$ onto the 2D camera space as shown in Figure~\ref{fig:proj_process}, computing a set of camera foreground conditions $\mathcal{M} = \{ \textbf{b}^v, \textbf{l}^v, \textbf{h}^v \}$ and background conditions $ \mathcal{R} = \{ \textbf{R}^v \} $ of all view, which details see Sec.~\ref{sec:Method_controller}.
Then we encode camera foreground and background conditions into a set of camera foreground and background embedding $\mathcal{F} = \{\textbf{F}^v \in \mathbb{R}^{N \times C}  \}$ and $\mathcal{B} = \{\textbf{B}^v \in \mathbb{R}^{(H \times W) \times C}  \}$, where $N$ denotes the number of bounding boxes in each view.  
Through the \textbf{Controller}, each perspective can obtain semantic control information from the foreground and the background embedding of the corresponding camera view.
This process results in the generation of geometry-consistent street-view latent features $ \mathcal{G}_t = \{\textbf{G}_t^v \in \mathbb{R}^{ H \times W \times C}  \}$. Next, the geometry-consistent features $ \mathcal{G}_t $ are fed into \textbf{Coordinator}. The Coordinator employs a novel cross-view-cross-element attention mechanism to enhance adjacent views' consistency, yielding the appearance-consistent street-view latent features $ \mathcal{A}_t = \{\textbf{A}_t^v \in \mathbb{R}^{ H \times W \times C }  \}$.

Then BEVControl employs the cross-attention mechanism of the diffusion model to handle the text prompt, allowing us to control the generated images' environmental factors (e.g., weather and lighting conditions). Then BEVControl repeats the execution of this UNet formed by stacking these three blocks $T$ times. Eventually, the output is the generated street-view images $ \mathcal{Z}_0 = \{ \textbf{Z}_0^v \in \mathbb{R}^{ H \times W \times C}  \} $, which are geometry-consistent, appearance-consistent and caption-aligned.

\subsection{Controller}
\label{sec:Method_controller}
% Formulations and description.
Based on the internal and external parameters of different cameras, we project the foreground and background classes of the BEV sketch $\mathcal{S}$ onto the corresponding pixel coordinate system to obtain the camera foreground conditions $\mathcal{M}$ and background conditions $\mathcal{R}$.

We define the camera foreground conditions as $\mathcal{M} = \{ \textbf{b}^v, \textbf{l}^v, \textbf{h}^v \}$, where $ \textbf{b}^v\in{[0,1]}^{N \times 4}, \textbf{l}^v $ and $ \textbf{h}^v\in{[-180, 180)}^{N \times 1} $ denotes the normalized pixel coordinates of the upper left and lower right corners, text label and heading degree of $N$ boxes in the current perspective. 
The camera background conditions $ \mathcal{R} =  \{\textbf{R}^v \in \mathbb{R}^{ H \times W \times 3 }  \}$, which are spatially aligned with the authentic camera images, representing the trend of the road.
Then we extract the camera foreground embedding $\mathcal{F} = \{\textbf{F}^v \in \mathbb{R}^{N \times C}  \}$ and background embedding $\mathcal{B} = \{\textbf{B}^v \in \mathbb{R}^{(H \times W) \times C}  \}$ as below: 
\begin{equation}
    \begin{aligned}
&   \begin{array}{c} 
        % w/ view prefix 
        {\textbf{F}^v} = \text{linear}( \text{fe}({\textbf{b}^v }) + \text{cte}({\textbf{l}^v}) + \text{fe}({\textbf{h}^v }) ) \vspace{2ex}, \\
        {\textbf{B}^v} = \text{cnn}( \textbf{R}^v ),
        % w/o view prefix 
        % & {\bf{F}} = \text{linear}( f({\textbf{b} }) + clip({\textbf{l}}) + f({\textbf{o} }) ) ,\\
        % & {\bf{B}} = \text{cnn}( \mathcal{B} ),
    \end{array} \\
    \end{aligned}
    \label{con:foreground_embedder}
\end{equation}
where $\text{fe}$ denotes Fourier Embedder~\cite{mildenhall2021nerf}, $\text{cte}$ denotes CLIP Text Encoder~\cite{radford2021_clip}, and $\text{cnn}$ denotes a pre-trained CNN network~\cite{liu2022_convnet}. 
Based on the existing extensive pre-trained diffusion model~\cite{rombach2022_ldm, li2023_gligen}, we inject the foreground and background embedding $\mathcal{F}$ and $\mathcal{B}$ by adding two trainable self-attention layer to the UNet architecture. The calculation formula is shown below: 
\begin{equation}
    \begin{aligned}
        % w/ view prefix 
        & \textbf{G}_t^v = \textbf{Z}_t^v + \alpha \cdot \text{sa}([\textbf{Z}_t^v, {\textbf{F}^v}]) + \beta \cdot \text{sa}([\textbf{Z}_t^v, {\textbf{B}^v}]), 
        % % w/o view prefix 
        % & \textbf{G}_t = \textbf{Z}_t + \alpha \cdot \text{sa}([\textbf{Z}_t, {\textbf{F}}]) + \beta \cdot \text{sa}([\textbf{Z}_t, {\textbf{B}}]), 
    \end{aligned}
    \label{con:controller}
\end{equation}
where $[ \cdot ]$ denotes the concatenation operation. $\text{sa}$ denotes the self-attention block. $\alpha$ and $\beta$ are trainable parameters initialized to 0. 
The introduced self-attention layer can effectively find the mapping relationship between visual latent features and various camera condition embedding. Therefore, the controller can utilize spatial hints to output a set of latent features $ \mathcal{G}_t = \{\textbf{G}_t^v \in \mathbb{R}^{ H \times W \times C}  \}$, which geometry is consistent with the corresponding camera foreground and background conditions.

\subsection{Coordinator}
% Formulations and description.
Taking $\mathcal{G}_t = \{\textbf{G}_t^v \} $ as input, we employ the Coordinator to enhance the consistency of different views and make them look like somebody capture them from the same scene.

Specifically, we propose a novel cross-view-cross-element attention mechanism that enables context interaction between the different views. Sufficient context interaction makes the semantics of visual elements in various perspectives uniform.
According to the characteristics of ring-shaped cameras, each camera has the highest correlation with its adjacent cameras. Therefore, we carefully design each view to interact only with the contextual information of adjacent views, reducing the demand for computing resources. In particular, we let all camera views learn the context of their adjacent views in parallel. 
The context comprises two layers of information: global level and local level. The global level represents the entire latent feature of the previous perspective, while the local level refers to the specific element feature. 
Taking adjacent view $v$ and $u$ as an example, the learning context is $ \textbf{k}, \textbf{v}$ as shown below for view $v$: 
\begin{equation}
    \begin{aligned}
        % \left\{
        % & 
        \begin{array}{c} 
            {\bf q} = \text{linear}({\textbf{G}_t^v})\vspace{2ex},  \\
            {\bf k} = \text{linear}({[\textbf{G}_t^u, \textbf{F}^u, \textbf{B}^u ]}) \vspace{2ex}, \\
            {\bf v} =\text{linear}({[\textbf{G}_t^u, \textbf{F}^u, \textbf{B}^u ]}) ,
        \end{array}
        % \right.
    \end{aligned}
\end{equation}
where $\text{linear}$ modules above are independent of each other, and we set $u=1, v=V$ or $v=u+1$ to enforce $u $ and $v$ as the adjacent views. The context interaction process of adjacent views is formulated as:
\begin{equation}
\begin{aligned}
% \left\{
&
    \begin{array}{c} 
    % {\bf a} =   \vspace{2ex}, \\
    \textbf{A}_t^v =\textbf{G}_t^v + {\bf v}^{\top} \cdot \text{softmax}({\bf k} \cdot {\bf q}^{\top}).
    \end{array} \\
% \right.
\end{aligned}
\end{equation}
We perform the above operation on all views in parallel, resulting in a set of street-view latent feature $\mathcal{A}_t =  \{\textbf{A}_t^v \in \mathbb{R}^{ H \times W \times C }  \}$ . By interacting between the global and local levels, global information, such as environmental conditions and weather, and local information, such as object appearance and identity, can be transmitted from the previous to the following perspective. Thus the cross-view-cross-element attention effectively improves the appearance consistency of the street-view images.

\subsection{Training Objective}
By repeatedly applying the above UNet ${\epsilon}_\theta$ in the latent space, we can obtain street-view images with gradually reduced noise. By adding $t$ step noise $\epsilon$ to the original clear images $\mathcal{Z}_0$, we obtain a noisy version $\mathcal{Z}_t$ of the images. We train ${\epsilon}_\theta$ to predict the noise we added. Following the training objective of the original LDM~\cite{rombach2022_ldm}, we finetune the pre-trained diffusion model~\cite{li2023_gligen} to adapt to new conditions $c$ (e.g. BEV sketch and text prompt):
\begin{align}
\label{eq:ldm_loss}
% \small
\min_{\theta} \mathcal{L} = \mathbb{E}_{\mathcal{Z}_0, \epsilon \sim  \mathcal{N}(\mathbf{0}, \mathbf{I}), t, c} \big[ \|  \epsilon -   {\epsilon}_\theta (\mathcal{Z}_t, t, c) \|^2_2 \big],
\end{align}
where time step $t$ is uniformly sampled from $[1, T]$, and $\theta$ refers to the newly added layer in the UNet. We only train the newly introduced layer while freezing the layers of the original diffusion model. This approach reduces memory consumption and avoids knowledge forgetting and model collapsing. 
\section{Evaluation Metrics for Content Controlling}

% Formulations and figures of how to compute the scores with respect to these metrics (how to compute OGS).
\begin{figure}[t!]
\centering
\includegraphics[width=\linewidth]{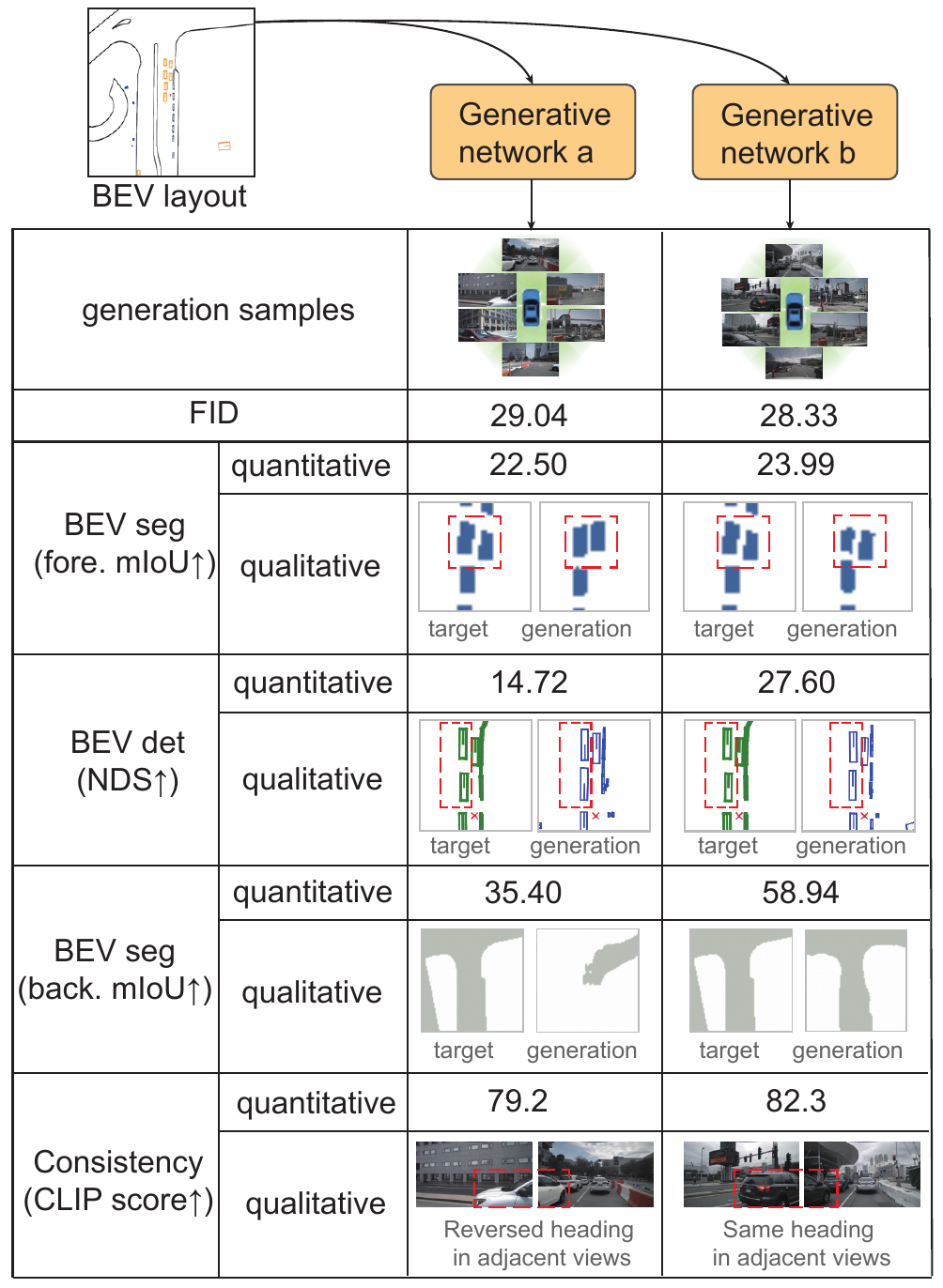}
\vspace{-0.1in}
\caption{Comparison of detail evaluation metrics.}
\label{fig:eval_metric}
\vspace{-0.15in}
\end{figure}

Recent street view image generation works~\cite{swerdlow2023_bevgen} only evaluate the generation quality based on scene-level metrics such as FID, vehicle mIoU, and road mIoU. However, we found that using only these metrics cannot evaluate the true generation ability of the generative network. As shown in the Figure~\ref{fig:eval_metric}, the reported qualitative and quantitative results simultaneously indicate that several sets of generated street view images with similar FID scores have vastly different fine-grained control abilities over foreground and background. Therefore in this section, we introduce the evaluation metrics in our experiment for measuring the controlling power of the generative network.

\vspace{0.05in}
\noindent{\bf Evaluation Metrics for Realism, Diversity, and Consistency~~}
Given the street-view images $\{{\bf X}'_{v}~|~v=1,...,V\}$ from $V$ perspectives output by the image decoder of CLIP~\cite{radford2021_clip}, we use the Frechet Inception Distance (FID)~\cite{heusel2017_fid} to measure the realism and diversity of the generated street-view images. We compute the FID between the latent features of the generated and real images, which capture the same perspective's foreground and background contents. Here, we employ the Inception-V3 network~\cite{szegedy2016_inception} to extract the latent features of the generated and real images. We compute the average FID score $S_{\text{FID}} \in \mathbb{R}$ as:
\begin{equation}
\begin{aligned}
S_{\text{FID}} = \frac{1}{V} \sum^V_{v=1} \text{fid}(\sigma({\bf X}_{v}), \sigma({\bf X}'_{v})).
\label{eval:FID}
\end{aligned}
\end{equation}
$\{{\bf X}_{v} \in \mathbb{R}^{H \times W \times 3}~|~v=1,...,V\}$ are the real images. We denote $\sigma$ as the Inception-V3 network. $\sigma({\bf X}_{v}), \sigma({\bf X}'_{v}) \in \mathbb{R}^{C}$ are the latent features of the $v^{th}$ perspective's generated and real images. A lower FID score $S_{\text{FID}}$ means that the generated contents are more realistic and diverse.

To evaluate the visual consistency between the generated street-view images, we compute the CLIP score~\cite{radford2021_clip}, based on the latent features of the overlap between the adjacent perspectives of the generated street-view images. We calculate the CLIP score $S_{\text{CLIP}} \in \mathbb{R}$ as:
\begin{equation}
\begin{aligned}
& S_{\text{CLIP}} = \frac{1}{V} \sum_{u,v} \text{clip}(\psi({\bf X}'_u), \psi({\bf X}'_v)),\\
&~~~~~s.t.,~~u=1, v=V~\text{or}~v=u+1,
\label{eval:CLIP}
\end{aligned}
\end{equation}
where ${\bf X}'_u, {\bf X}'_v$ are the generated street-view images of the adjacent perspectives. We denote $\psi({\bf X}'_{v-1}), \psi({\bf X}'_{v}) \in \mathbb{R}^{C}$ as latent features of the overlap between ${\bf X}'_u, {\bf X}'_v$. A higher CLIP score means satisfactory visual consistency.

\vspace{0.05in}
\noindent{\bf Evaluation Metrics for Foreground and Background Controlling~~}
We employ the official detection metrics of the nuScenes dataset, i.e., the mean average precision (mAP), the nuScenes detection score (NDS), and the mean average orientation error (mAOE), for measuring the foreground controlling score. We denote the scores of mAP, NDS, and mAOE as $S_{\text{AP}}$, $S_{\text{NDS}}$, and $S_{\text{AOE}}$. Specifically, based on the generated street-view images $\{{\bf X}'_{v}~|~v=1,...,V\}$, we use BEVFormer~\cite{li2022_bevformer} trained on the nuScenes dataset to detect the foreground objects on the BEV layouts. We achieve the scores $S_{\text{AP}}$, $S_{\text{NDS}}$, and $S_{\text{AOE}}$ of foreground object detection by comparing the detection results with the BEV layout used for generating the street-view images. We use CVT~\cite{zhou2022_CVT} trained on the nuScenes dataset to segment the foreground on the BEV layouts and report the foreground mean intersection-over-union (fIoU) performance denoted as $S_{\text{fIoU}}$. To evaluate the background controlling power of the generative network, we employ CVT to segment the background contents on the BEV layouts and report the performance in terms of the background mean intersection-over-union (bIoU) denoted as $S_{\text{bIoU}}$.

We remark that higher scores of $S_{\text{AP}}$, $S_{\text{NDS}}$, $S_{\text{fIoU}}$, $S_{\text{bIoU}}$, and a lower score of $S_{\text{AOE}}$ mean a good controlling power of the generative network, which produces the foreground and background contents corresponding to the ground-truth annotations in the BEV layouts.

\vspace{0.05in}
\noindent{\bf Overall Evaluation Metric~~}
We propose a combinatorial metric to summarize the above metrics that measure the controlling power of the generative network from separate aspects. We name this combinatorial metric as the overall controlling score (OCS) denoted $S_{\text{OCS}}$. We compute $S_{\text{OCS}}$ as:
% \begin{equation}
% \begin{aligned}
% &~~~~~~~~~~~~~~~~~S_{\text{OCS}} = \sum_r \exp(\max(1-r,0)),\\
% & s.t.,r \in \{\frac{U_{\text{FID}}}{S_{\text{FID}}}, \frac{S_{\text{CLIP}}}{U_{\text{CLIP}}}, \frac{S_{\text{NDS}}}{U_{\text{NDS}}}, \frac{S_{\text{fIoU}}}{U_{\text{fIoU}}}, \frac{S_{\text{bIoU}}}{U_{\text{bIoU}}}\}.
% \label{eval:OCS}
% \end{aligned}
% \end{equation}
\begin{equation}
\begin{aligned}
S_{\text{OCS}} = \frac{U_{\text{FID}}}{S_{\text{FID}}}+\frac{S_{\text{CLIP}}}{U_{\text{CLIP}}}+\frac{S_{\text{NDS}}}{U_{\text{NDS}}}+\frac{S_{\text{fIoU}}}{U_{\text{fIoU}}}+\frac{S_{\text{bIoU}}}{U_{\text{bIoU}}}.
\label{eval:OCS}
\end{aligned}
\end{equation}
The scores $\{S_{\text{FID}}, S_{\text{CLIP}}, S_{\text{NDS}}, S_{\text{fIoU}}, S_{\text{bIoU}}\}$ are achieved by using BEVFormer to detect and CVT to segment street-view contents on the BEV layouts, according to the generated images. We define another set of reference scores $\{U_{\text{FID}}, U_{\text{CLIP}}, U_{\text{NDS}}, U_{\text{fIoU}}, U_{\text{bIoU}}\}$, which are detection and segmentation performances based on the authentic images. A high score of $S_{\text{OCS}}$ means the entire controlling power is strong.
\section{Experimental Results}

%\subsection{Implementation Details}
%leave it to supplementary file?

\subsection{Dataset}

%\vspace{0.01in}
%\noindent{\bf Datasets~~}

%\vspace{0.05in}
%\noindent{\bf Data Processing~~}

We use the public nuScenes dataset~\cite{caesar2020_nuscenes} to examine the effectiveness of our method. nuScenes contains 1,000 examples of street-view scenes. There are 700/150/150 training/validation/testing examples. Each example records about 40 frames of BEV layouts. Each frame of the BEV layout is associated with six street-view RGB images, which are captured by an ego vehicle's side, front, and back cameras. We follow the convention~\cite{swerdlow2023_bevgen} to sample 600 frames of BEV layouts from the validation set, forming a validation subset to evaluate our method.

The objects in each BEV layout are annotated as the foreground and background. For object detection, the foreground includes ten categories (i.e., car, bus, truck, trailer, motorcycle, bicycle, construction vehicle, pedestrian, barrier, and traffic cone), while the background is the road. For object segmentation, the categories of car, bus, truck, trailer, motorcycle, bicycle, and construction vehicle are merged into the vehicle category. Thus, each BEV layout contains the binary categories of vehicle and road.

\subsection{Visual Element Control}

In Table~\ref{tab:control}, we compare BEVControl with the recent methods~\cite{swerdlow2023_bevgen, li2023_gligen, zheng2023_layoutdiffusion}, which can also generate the street-view images based on the BEV layout. Given a BEV layout, each method generates a set of street-view images from 6 perspectives. The results in Table~\ref{tab:control} measure the quality of the generated images and the controlling power of the compared methods on the background/foreground objects.
We also report the performance improvement of BEVControl relative to the GLIGEN~\cite{li2023_gligen} in the last row.
BEVControl achieves a higher OGS than other methods (see the right-most column of Table~\ref{tab:control}). Below, we evaluate the detailed performances of controlling various visual elements.

% It should be noted that a generative model with a strong controlling power should produce the images with realism and diversity. 

\setlength{\tabcolsep}{3.7pt}
\renewcommand{\arraystretch}{1.2}
\begin{table*}
\centering
\small
\begin{threeparttable}
\begin{tabular}{c|c|c|ccc|c|c|c}
\hline
 \multirow{3}{*}{\textbf{Method}} & \multicolumn{1}{c|}{\textbf{Real \& Diverse}} & \multicolumn{1}{c|}{\textbf{Consistency}}  &  \multicolumn{4}{c|}{\textbf{Foreground Control}}  & \multicolumn{1}{c|}{\textbf{Background Control}} & \multirow{3}{*}{$ S_{\text{OCS}} \uparrow $} \\
 \cline{2-8}
 
 &\multirow{2}{*}{$S_{\text{FID}}\downarrow$} &\multirow{2}{*}{$ S_{\text{CLIP}} \uparrow $} & \multicolumn{3}{c|}{\textbf{Detection}}  & \multicolumn{1}{c|}{\textbf{Segmentation}}  & \multicolumn{1}{c|}{\textbf{Segmentation}} \\
 \cline{4-8}

& \multicolumn{1}{c|}{\textbf{}} & \multicolumn{1}{c|}{\textbf{}}
& \multicolumn{1}{c}{$ S_{\text{AP}} \uparrow $} & \multicolumn{1}{c}{$ S_{\text{NDS}} \uparrow $} 
& \multicolumn{1}{c|}{$ S_{\text{AOE}} \downarrow $}  & \multicolumn{1}{c|}{$ S_{\text{fIoU}} \uparrow $}
& \multicolumn{1}{c|}{$ S_{\text{bIoU}} \uparrow $}
 \\ \hline
 Reference-score  &0.01 &87.96 & 36.04 &44.10 &0.42 & 34.83  &74.33 & 5.00 \\
\hline
BEVGen~\cite{swerdlow2023_bevgen}  &25.54 &- &- &- &- & 5.89 &50.20 & -\\
LayoutDiffusion~\cite{zheng2023_layoutdiffusion}  &29.64  &79.80 &3.68  &14.68  &1.31  &15.51  &35.31  &2.16  \\
GLIGEN~\cite{li2023_gligen}  &31.34   &78.80 &15.42 &22.35 &1.22 & 22.02 & 38.12 &2.55 \\ \hline
\multirow{2}{*}{\textbf{BEVControl}}  &\textbf{24.85}    &\textbf{82.70}   &\textbf{19.64} &\textbf{28.68} &\textbf{0.78} & \textbf{26.80} &\textbf{60.80} &\textbf{3.18}  \\ 
 
  &($\downarrow$ \textbf{6.49}) &($\uparrow$ \textbf{3.9}) &($\uparrow$ \textbf{4.22}) &($\uparrow$ \textbf{6.33}) &($\downarrow$ \textbf{0.44}) &($\uparrow$ \textbf{4.78}) &($\uparrow$ \textbf{22.68}) &($\uparrow$ \textbf{0.63}) \\
  \hline
\end{tabular}
\end{threeparttable}
\caption{We compare BEVControl with state-of-the-art methods on the validation subset of nuScenes. The results measure the controlling power of different methods. $\downarrow$/$\uparrow$ means a smaller/larger value of the metric represents a better performance.}
\label{tab:control}
\end{table*}

\begin{figure*}[t!]
\centering
\includegraphics[width=\linewidth]{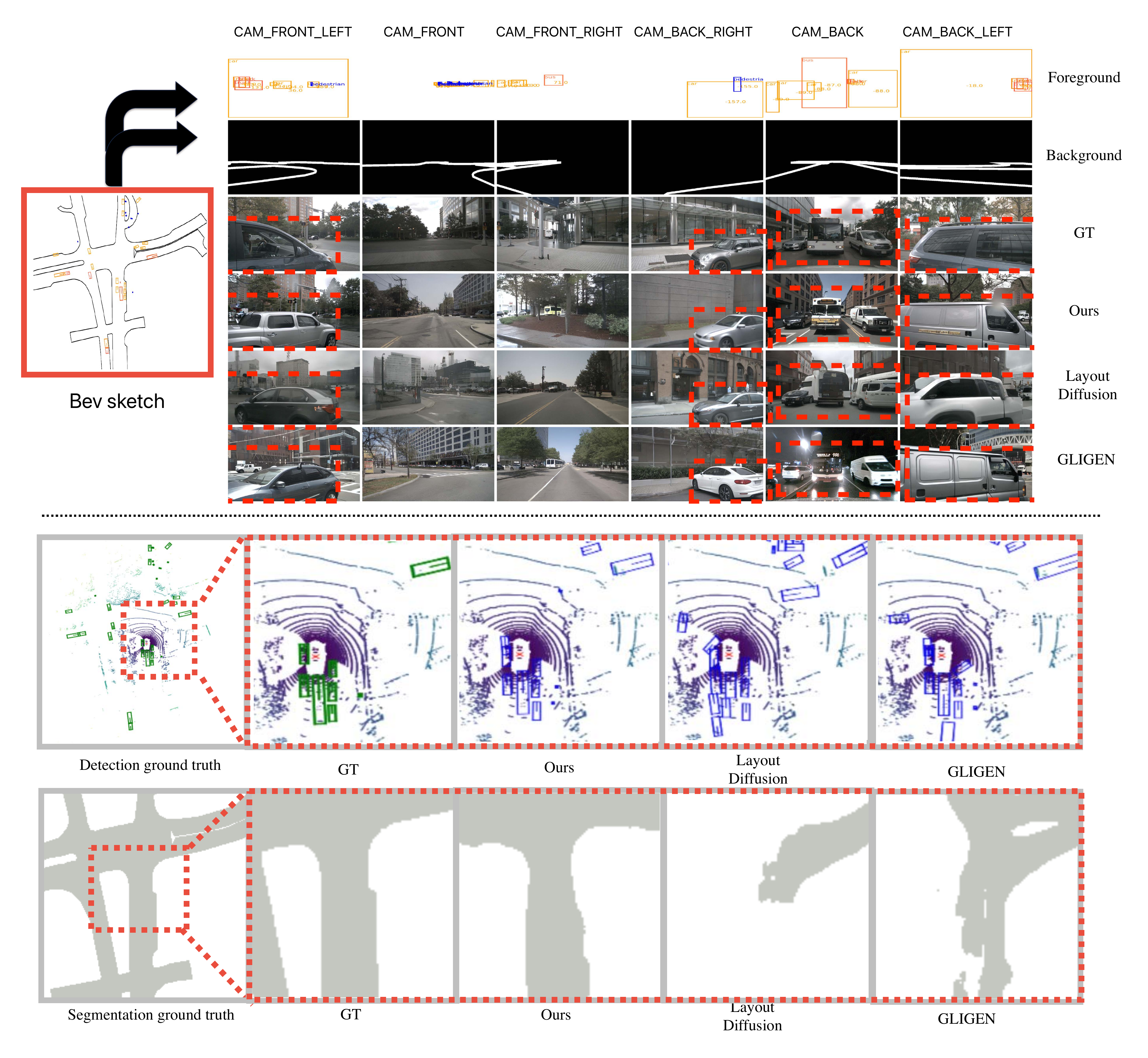}
\vspace{-0.1in}
\caption{The visualization of foreground controlling generation. Compared to other methods, ours can generate objects that correspond more closely to the bounding box sketch conditions, especially the accurate orientation.}
\label{fig:val_foreground_control_visual}
\vspace{0.2in}
\end{figure*}

\begin{figure*}[t!]
\centering
\includegraphics[width=\linewidth]{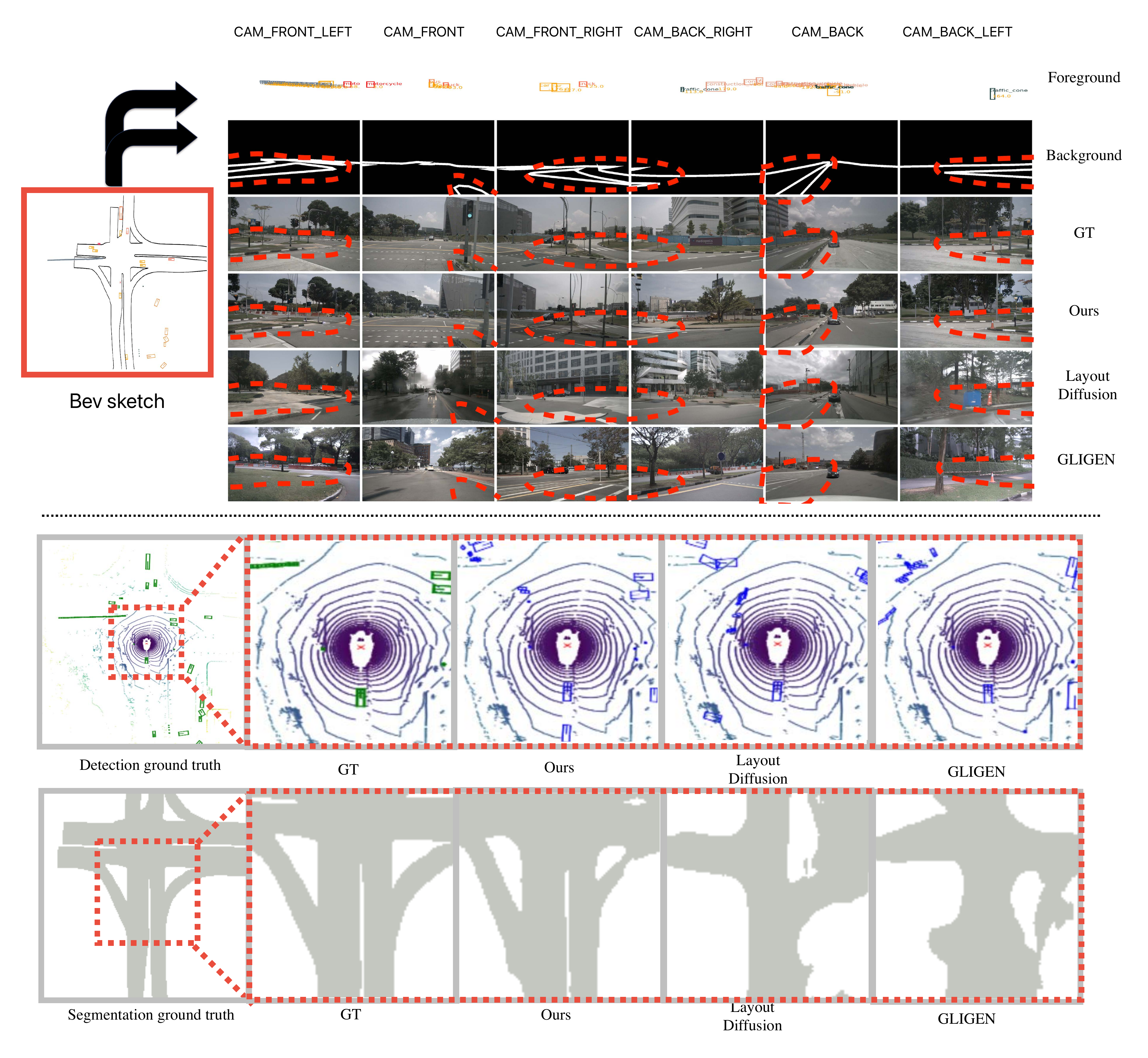}
\vspace{-0.1in}
\caption{The visualization of background controlling generation. Compared to other methods, ours can generate street views that correspond more closely to the road sketch conditions.}
\label{fig:val_background_control_visual}
\vspace{0.2in}
\end{figure*}

\vspace{0.05in}
\noindent{\bf Realism and Diversity~~} 
In Table~\ref{tab:control} ``Real \& Diverse", we measure the realism and diversity of the image data generated by different methods in terms of the Frechet Inception Distance (FID). BEVControl achieves 24.85 FID, outperforming other methods. We also compare the street-view images generated differently in Figure~\ref{fig:val_foreground_control_visual} and~\ref{fig:val_background_control_visual}, where BEVControl produces a higher image quality than the compared methods.

\vspace{0.05in}
\noindent{\bf Foreground Control~~}
% In Table~\ref{tab:control} ``Foreground Control", we examine the controlling power of different methods on the foreground objects. In this examination, we use BEVFormer~\cite{li2022_bevformer} trained on the nuScenes dataset to detect the ten categories of foreground objects in the BEV layouts based on the street-view images generated by different methods. We report the performance of foreground object detection in terms of mAP, NDS, and mAOE. Based on the generated street-view images, we use BEVFormer to segment the foreground (i.e., vehicle) on the BEV layouts and report the performance in terms of mIoU. Better performances of detection and segmentation mean the generated foreground objects correspond to the ground-truth annotations in the BEV layouts that play as the hints for the generation. In the first row of Table~\ref{tab:control} ``Reference-score", we report the detection and segmentation performances of BEVFormer on the original validation subset. These results can be regarded as the Reference-score performance of BEVFormer for measuring the controlling power of BEVControl.
%
In Table~\ref{tab:control} ``Foreground Control", we examine the controlling power of different methods on the foreground objects. In this examination, we use BEVFormer~\cite{li2022_bevformer} to detect the ten categories of foreground objects on the BEV layouts, reporting the performance in terms of mAP, NDS, and mAOE. We use CVT~\cite{zhou2022_CVT} to segment the foreground (i.e., vehicle) on the BEV layouts, whose performance is reported in terms of mIoU. In the first row of Table~\ref{tab:control} ``Reference-score", we report the detection performances of BEVFormer and segmentation performances of CVT on the original validation subset. These results can be regarded as the Reference-score performance of BEVFormer and CVT for measuring the controlling power of BEVControl.

\begin{figure*}[t!]
\centering
\includegraphics[width=\linewidth]{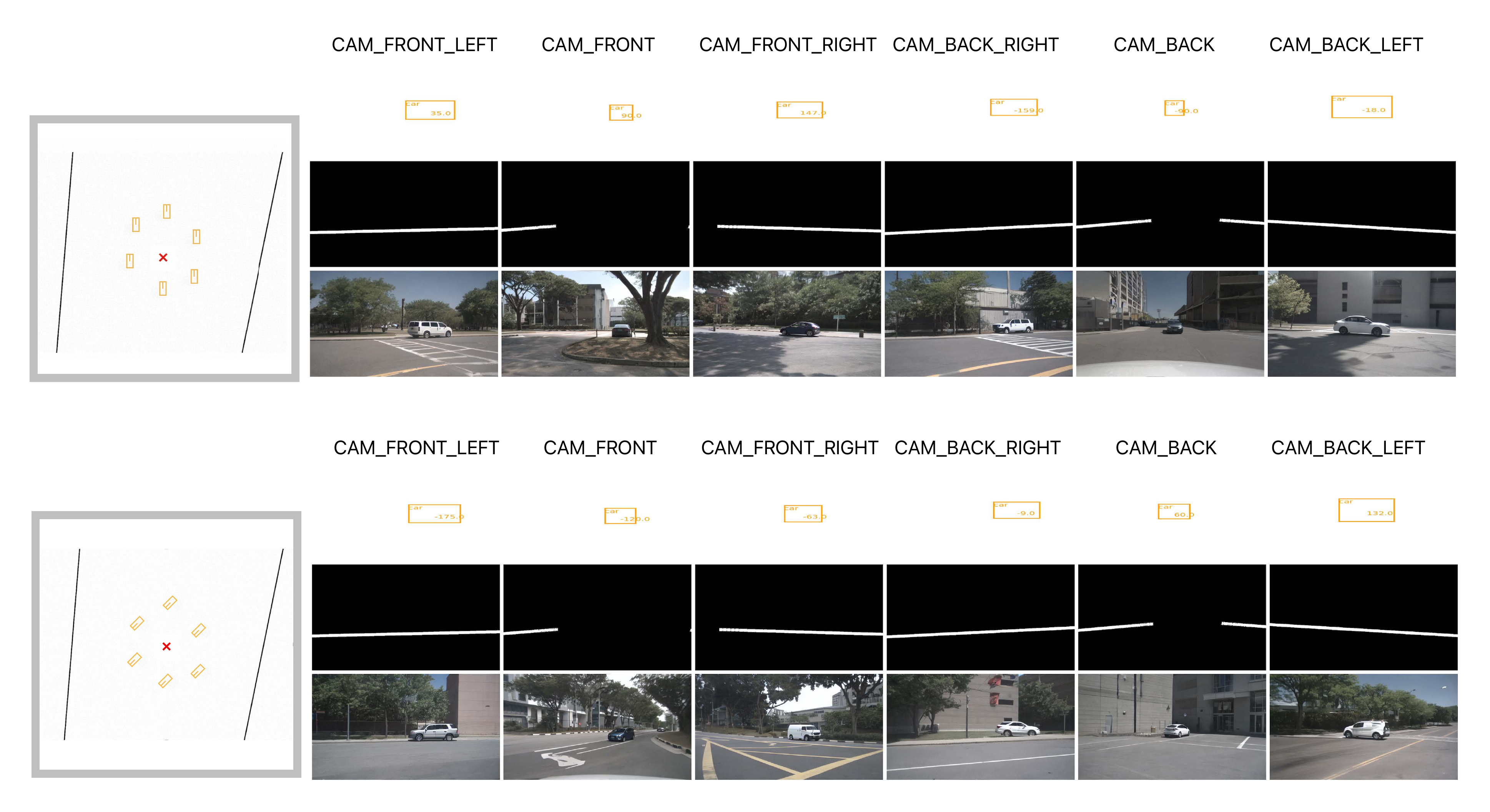}
\vspace{-0.1in}
\caption{The visualization of foreground controlling generation in various vehicle orientation . }
\label{fig:sketch_foreground_control_visual}
\vspace{0.2in}
\end{figure*}

Based on the data generated by BEVControl, BEVFormer and CVT achieve better detection and segmentation performances than other generative models. It demonstrats a more substantial controlling power of BEVControl on the foreground objects. We compare the generated foreground objects by different methods in Figure~\ref{fig:val_foreground_control_visual}, where BEVControl satisfactorily yields the foreground objects according to the ground-truth annotations. Furthermore, Figure~\ref{fig:sketch_foreground_control_visual} demonstrates the generation capability of BEVControl for user-drawn BEV sketches of different vehicle orientations.

\begin{figure*}[t!]
\centering
\includegraphics[width=\linewidth]{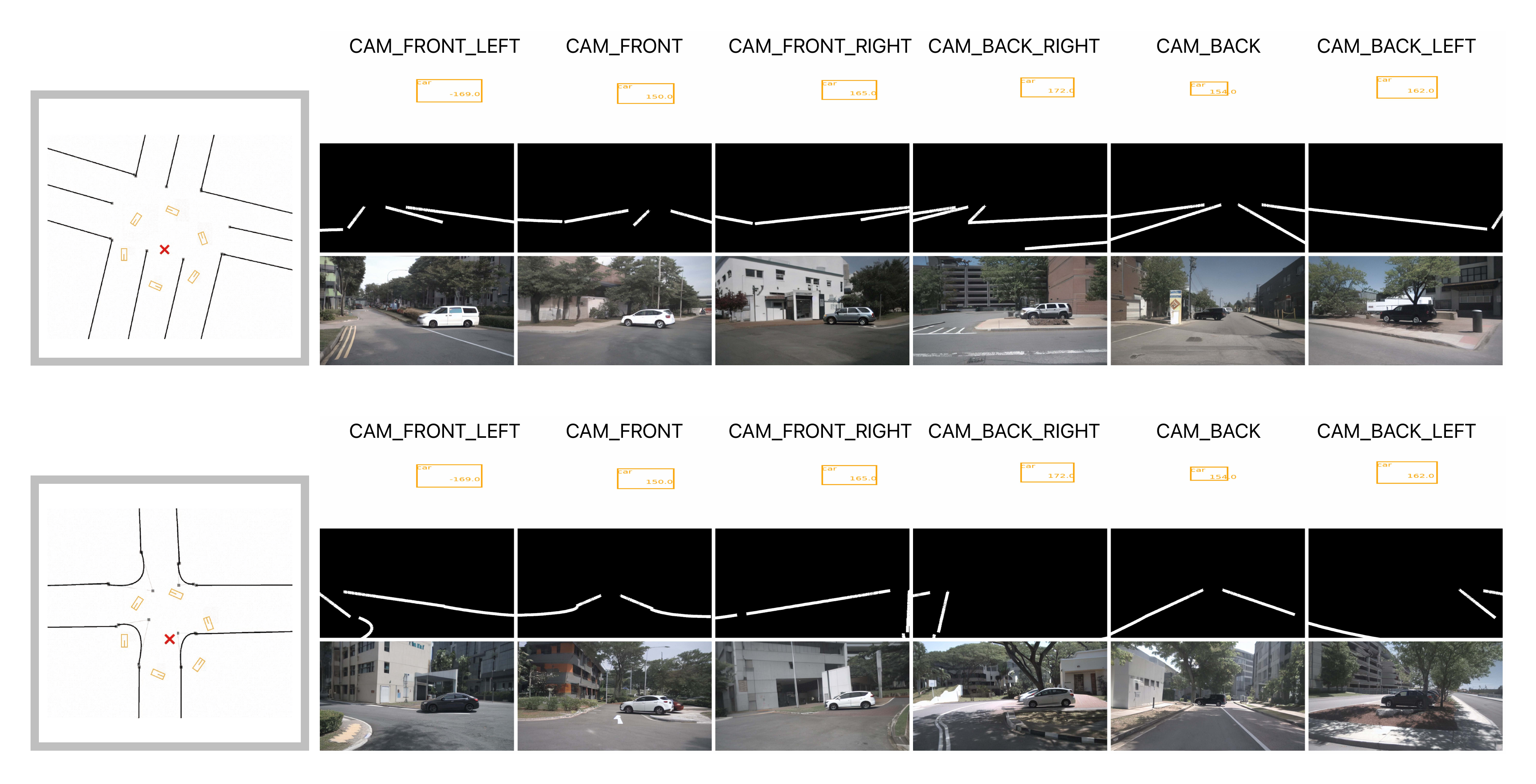}
\vspace{-0.1in}
\caption{The visualization of background controlling generation in various road sketch. }
\label{fig:sketch_background_control_visual}
\vspace{0.2in}
\end{figure*}

\vspace{0.05in}
\noindent{\bf Background Control~~}
In Table~\ref{tab:control} ``Background Control", we study the controlling power of different methods on the background objects. Again, we use the trained CVT, which uses the generated street-view images to segment the background (i.e., road) on the BEV layouts. We report the segmentation accuracy mIoU on the road category. We also compare the generated roads by different methods in Figure~\ref{fig:val_background_control_visual}. Compared to the street-view images generated by other methods, those generated by BEVControl lead to a better segmentation accuracy, which means a more substantial controlling power of BEVControl on the background objects.  Additionally, Figure~\ref{fig:sketch_background_control_visual} displays the generation capability of BEVControl for user-edited BEV sketches of different road traffic situations.

\subsection{Ablation Study on Controller}

% \setlength{\tabcolsep}{6pt}
%   \begin{table}[t!]
%     \centering
%     \begin{tabular}{c||ccc|c}
%     \hline
%     \textbf{Controller}  & \textbf{FID} $\downarrow$ & \textbf{FC} $\uparrow$ & \textbf{BC} $\uparrow$ &\textbf{OCS} $\uparrow$            \\ \hline \hline
%     Reference-score &0  &2  &1     &             \\\hline\hline
%     foreground      & \checkmark   &   &      &             \\\hline
%     background      & training   &   &      &             \\\hline\hline
%     both w/o separation &   &     &    &             \\\hline 
%     both w/ separation & \checkmark   &     &    &             \\\hline 
%     \end{tabular}
%     \caption{}
%     \label{tab:condition}
%   \end{table}
  
\setlength{\tabcolsep}{6.5pt}
\renewcommand{\arraystretch}{1.2}
\begin{table}
\centering
\small
\begin{threeparttable}
\begin{tabular}{c|cc|c|c}
\hline
 \multirow{2}{*}{\textbf{Controller}} & \multicolumn{2}{c|}{\textbf{FC} } & \multicolumn{1}{c|}{\textbf{BC} }  &  \multirow{2}{*}{$ S_{\text{OCS}} \uparrow $} \\
 \cline{2-4}
 
 &\multirow{1}{*}{$ S_{\text{NDS}} \uparrow $} &\multirow{1}{*}{$ S_{\text{fIoU}} \uparrow $} & \multirow{1}{*}{$ S_{\text{bIoU}} \uparrow $} \\ \hline 

Reference-score &44.10  &34.83  &74.33     &5.00             \\\hline
foreground      & 25.23   & 22.50   & 41.70      & 2.69            \\
background      & 3.70    & 3.53   & 49.71      & 1.74            \\\hline
both w/o separation &26.87   &23.78     &52.30    & 2.90            \\
both w/ separation & \textbf{28.68}   &\textbf{26.80}     &\textbf{60.80}    & \textbf{3.18}            \\\hline 

\end{tabular}
\end{threeparttable}
\caption{Different strategies of using the foreground and background hints for controlling the visual elements. The evaluation metrics (i.e., FID, scores of foreground and background control) reported in this table are the same to those in Table~\ref{tab:control}.}
\label{tab:condition}
\end{table}
  
The controller of BEVControl regards the bounding boxes and the road sketches as hints. The diffusion model uses these hints to generate the foreground and background objects in the street-view images. Here, we experiment with different strategies of using the foreground and background hints to examine their effect on controlling the visual elements in the generated street-view images. We report the quantitative results in Table~\ref{tab:condition}.

First, we use the foreground or background hint only, reporting the scores of FID, foreground control (FC), and background control (BC) in the second and third rows of Table~\ref{tab:condition}. Without foreground or background hint, we degrade the realism and diversity of the generated images, while the score of foreground or background control also decreases consequently. These results demonstrate the importance of using the foreground and background hints together.

Next, we compare various strategies for using both foreground and background hints. We evaluate an alternative method that employs the foreground and background hints without separately controlling the visual elements. We use a single attention layer to jointly embed the foreground and background hints into latent space. The controller uses the latent embedding of these hints for outputting an information-controlled feature map. The coordinator relies on the information-controlled feature map to generate the street-view images. We report the foreground or background control scores in the fourth row of Table~\ref{tab:condition}. Though the scores are higher than those achieved using the foreground or background hint alone, they still lag behind the results of the entire controller in the fifth row. The complete controller uses separate network streams to enable a more focused control of the foreground and background objects.

\subsection{Ablation Study on Coordinator}

The coordinator utilizes the CVCE attention to enhance the visual consistency between the generated street-view images. Here, we use the street-view images from various perspectives to compute the CLIP score for measuring visual consistency. In Table~\ref{tab:coordinator}, we compare the CLIP scores for the street-view images generated by different alternatives.

We remove the coordinator from BEVControl, which has the controller alone. This alternative significantly degrades the visual consistency of the generated street-view images (see the CLIP score in the second row of Table~\ref{tab:coordinator}). We improve the CLIP score by adding the coordinator with the cross-view attention but without the cross-element attention (see the third row). This result demonstrates the positive impact of cross-view attention on visual consistency.

\setlength{\tabcolsep}{20pt}
  \begin{table}[]
    \centering
    \begin{tabular}{c|c}
    \hline
\textbf{Coordinator}  & $ S_{\text{CLIP}} \uparrow $         \\ \hline 
    Reference-score         &87.96              \\ \hline
    w/o coordinator     &79.50              \\ 
    w/ CV, w/o CE       &82.30              \\ 
    % arxiv_v2
    % \hline
    % w/ CVCE, w/o foreground &           \\ 
    % w/ CVCE, w/o background &           \\ 
    w/ CVCE   &\textbf{82.70}           \\ \hline
    \end{tabular}
    \caption{Different strategies of using the coordinator for yielding the street-view images from different perspectives. We report the results as CLIP scores, which measure the visual consistency of the street-view images.}
    \label{tab:coordinator}
    \end{table}

\subsection{Data Augmentation for Object Detection}

Based on each BEV layout from the training set of the nuScenes dataset, we again employ BEVControl to generate a set of street-view images. Note that the BEV layout and the generated street-view images can be used together for augmenting the training set of nuScenes. We use the generated data for training BEVFormer for object detection on the BEV layout from the train  subset. We report the performances of object detection in Table~\ref{tab:data_aug}. Compared to the BEVFormer trained without data augmentation (see the first row), the counterpart with data augmentation yields better results (see the second row).

% arxiv_v2
% We also experiment with jittering the foreground objects in the BEV layouts. This jittering is done by randomly moving the vehicles in the BEV layouts (see the examples shown in Figure~\ref{?}). We eliminate those vehicles that are out of the road boundaries. The jittering produces new BEV layouts for generating the street-view images for data augmentation. With jittering,  BEVControl generates more street-view images for training BEVFormer, further improving detection performances (see the third row).

\setlength{\tabcolsep}{5.0pt}
  \begin{table}[h!]
    \centering
    \begin{tabular}{c|ccc}
    \hline
    \textbf{Method}  & $ S_{\text{AP}} \uparrow $ & $ S_{\text{NDS}} \uparrow $  & $ S_{\text{AOE}} \downarrow $        \\ \hline 
    w/o augmentation        &37.00   &47.90 &0.66             \\
    w/ augmentation        &\textbf{38.96}   &\textbf{49.19} &\textbf{0.42}               \\ 
    % arxiv_v2
    % w/ augmentation + jittering        &   &  &              \\ 
    \hline
    \end{tabular}
    % arxiv_v2
    % \caption{Different strategies of using the generated street-view images for augmenting the training data. We report the detection performances of BEVFormer on the validation set.}
    \caption{Application of using the generated street-view images for augmenting the training data. We report the detection performances of BEVFormer on the validation set.}
    \label{tab:data_aug}
  \end{table}

\section{Conclusion}
% \label{sec:conclusion}

Given the BEV layout as the hint, the most advanced generative networks can synthesize the street-view images with realistic and diverse appearances, thus enriching the data for training the BEV perception model and profiting the autonomous driving. This paper advocates the significance of strengthening the controlling power of the generative network for BEV perception. We propose a novel generative network, BEVControl, which relies on the sketches of the BEV layout to synthesize the background and foreground elements in the street-view images. By depending on more focused hints, BEVControl enables accurate control of the background and foreground elements, whose visual consistency across multiple perspectives is maintained by the cross-view-cross-element attention. Compared to the contemporary methods, a better controlling power allows BEVControl to yield richer data for BEV perception.

In future work, we will investigate how to better control more kinds of visual elements like lighting and weather in the generated images rather than the background and foreground only. In addition to generating street-view images, we will also study how to transfer the idea of BEVControl to create more general scenes.

{\small
\bibliographystyle{ieee_fullname}
\bibliography{11_references}
}

% \ifarxiv \clearpage \input{12_appendix} \fi

\end{document}

%% TITLE
\title{\paperTitle \\ Supplemental Material}
\author{\authorBlock}
\maketitle

\appendix
\section*{Appendix}
\label{sec:appendix}
In this supplemental material, we first provide more implementation and training details, and then present more results.

\section{Implementation and training details}
The number of view(e.g. $V$) is 6 in the NuScenes dataset~\cite{caesar2020_nuscenes} , arranged in clock-wise direction starting from cam\_front view. In the foreground condition, the orientation of bounding boxes is defined in the camera coordinate system as shown in Figure~\ref{fig:orientation_coords_sys}. 

% orientation_coords_sys
\begin{figure}[t!]
\centering
\includegraphics[width=\linewidth]{figs/method/box_orientation_sys_2.pdf}
\vspace{-0.1in}
\caption{Box orientation coordinate system. }
\label{fig:orientation_coords_sys}
\vspace{-0.15in}
\end{figure}

\begin{figure*}[t!]
\centering
\includegraphics[width=\linewidth]{figs/quanlitative_results/fg_bg_select/fg-bg-0046092508b14f40a86760d11f9896bb.pdf}
\vspace{-0.1in}
\caption{The visualization of foreground controlling generation. }
\label{fig:foreground_control_visual}
\vspace{0.2in}
\end{figure*}

\begin{figure*}[t!]
\centering
\includegraphics[width=\linewidth]{figs/quanlitative_results/fg_bg_select/fg-80e281bf369a4f849efaa5a9052cbd9b.pdf}
\vspace{-0.1in}
\caption{The visualization of foreground controlling generation. }
\label{fig:foreground_control_visual}
\vspace{0.2in}
\end{figure*}

\begin{figure*}[t!]
\centering
\includegraphics[width=\linewidth]{figs/quanlitative_results/fg_bg_select/bg-7ac3bb7fba5c4852a685555407cd10f1.pdf}
\vspace{-0.1in}
\caption{The visualization of background controlling generation.}
\label{fig:foreground_control_visual}
\vspace{0.2in}
\end{figure*}

\begin{figure*}[t!]
\centering
\includegraphics[width=\linewidth]{figs/quanlitative_results/fg_bg_select/bg-a4a9d61254d148fba35d53277b5246f8.pdf}
\vspace{-0.1in}
\caption{The visualization of background controlling generation.}
\label{fig:foreground_control_visual}
\vspace{0.2in}
\end{figure*}

\begin{figure*}[t!]
\centering
\includegraphics[width=\linewidth]{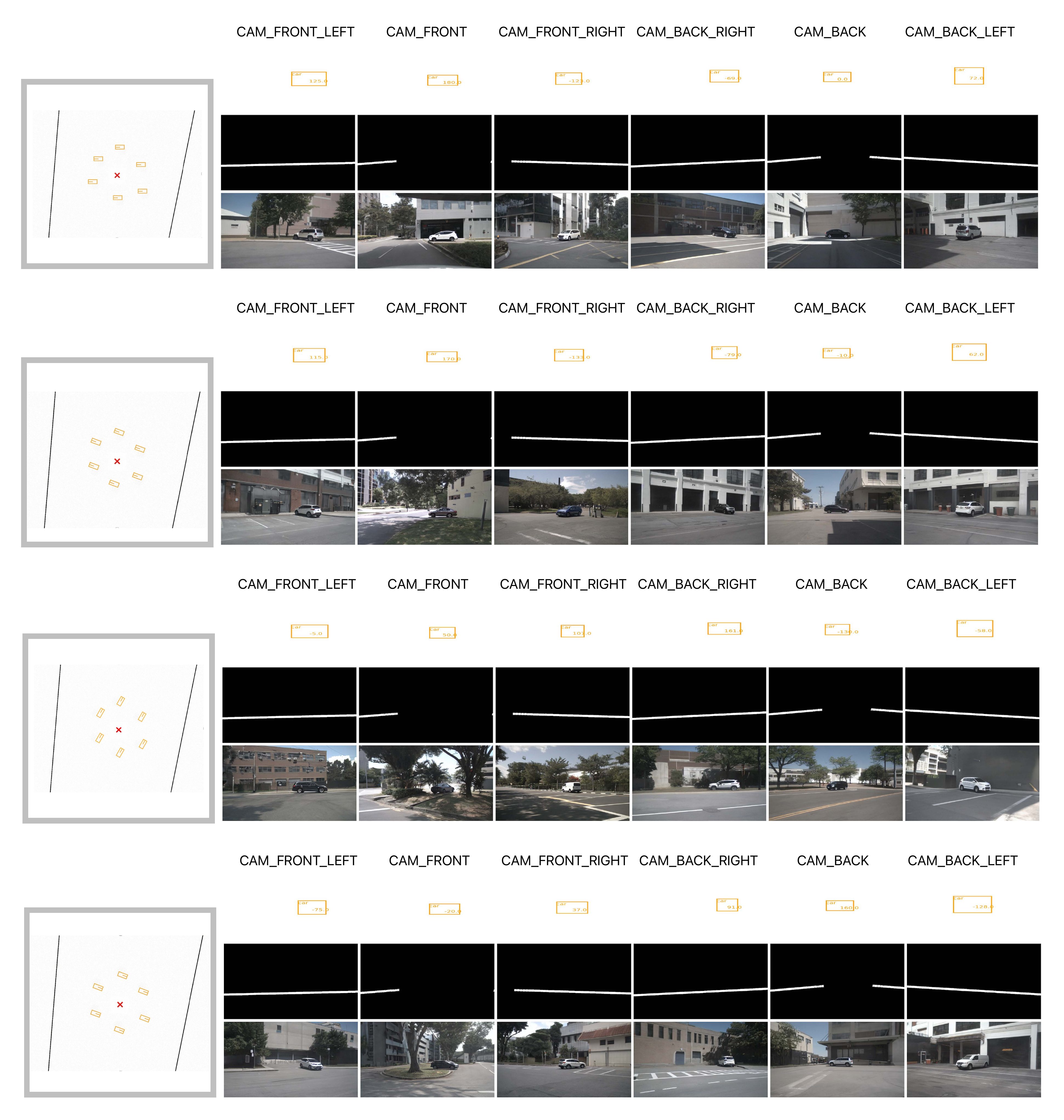}
\vspace{-0.1in}
\caption{The visualization of foreground controlling generation. }
\label{fig:foreground_control_visual}
\vspace{0.2in}
\end{figure*}

\begin{figure*}[t!]
\centering
\includegraphics[width=\linewidth]{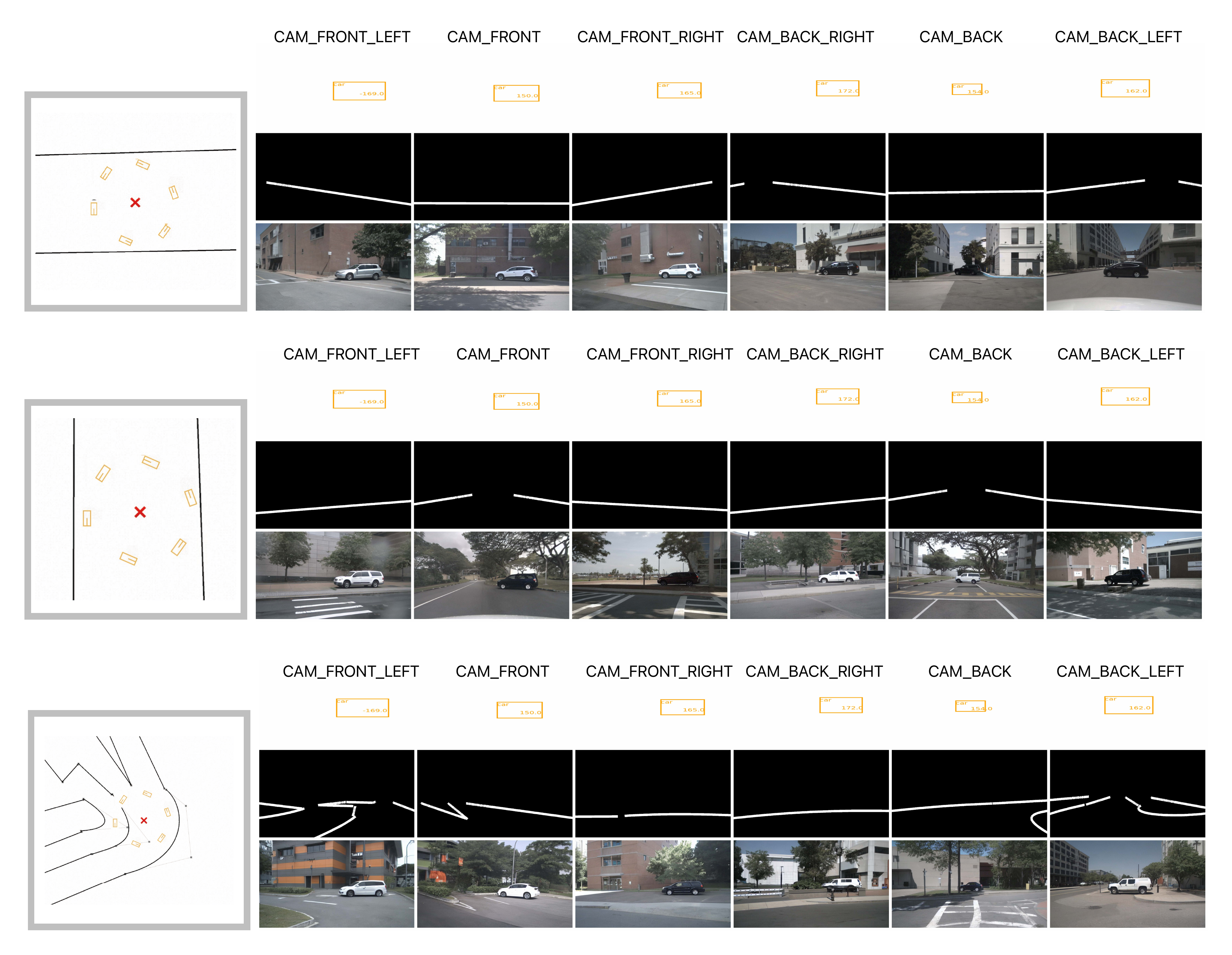}
\vspace{-0.1in}
\caption{The visualization of background controlling generation. }
\label{fig:foreground_control_visual}
\vspace{0.2in}
\end{figure*}

{\small
\bibliographystyle{ieee_fullname}
\bibliography{11_references}
}